\documentclass[letterpaper,10pt,conference]{ieeeconf}

\IEEEoverridecommandlockouts
\usepackage{amsmath,amssymb}
\usepackage{graphicx}
\usepackage{booktabs}
\usepackage{array}
\usepackage{cite}
\usepackage{url}
\usepackage{flushend}

\graphicspath{{figures/}}

\title{\LARGE \bf
From Learned-Mode AV--Traffic Pairing to Planner Decisions:\\
A Marginal-Preserving Study on Argoverse 2
}

\author{Jingyu Wang$^{1}$%
\thanks{$^{1}$Independent Researcher. Email: jwang.yalesds@gmail.com.}%
}

\begin{document}

\maketitle
\thispagestyle{empty}
\pagestyle{empty}

\begin{abstract}
Joint motion forecasts pair each autonomous-vehicle (AV) future with surrounding
traffic, but actor-level metrics do not show whether that structure matters to
a planner.  We study this question with a marginal-preserving product control that removes
AV--traffic pairing among the learned modes while retaining the fixed
constant-velocity pair and holding trajectories, actor-level
marginals before planner conditioning, candidates, the cost terms and weights,
and fallback fixed.
The intervention also changes candidate-conditioned concentration.

Across twelve runs on 1,400 held-out Argoverse 2 scenarios, the intervention
changes 3.0\% of route-level offline selections at $\tau=4$~m.  Control-minus-joint
recorded-trajectory regret is $-0.026$ and $-0.118$ at the two training sizes;
crossed and seed-$t$ intervals span zero.  At $\tau=1$~m, relative costs change
in 87.9\% of route evaluations and route-level offline selections in 8.1\%.
Before concentration matching, descriptive outcome estimates favor the
control.  Most of this gap disappears along an approximate
concentration-matching path; the remaining contrasts are $+0.112$ and $-0.047$, and
both crossed intervals span zero.  Actor-level forecast metrics remain
identical.  Pairing-strength and temperature sweeps show that the decision
contrast grows with pairing removal and sharper conditioning.  The intervention
changes planner decisions even though actor-level metrics remain unchanged.
The matching analysis, however, cannot separate any recorded-outcome
effect of learned-mode AV--traffic pairing from the accompanying change in conditioned
concentration.
\end{abstract}

\section{INTRODUCTION}

Forecast benchmarks commonly score actors separately using displacement
error~\cite{weng2023joint}.  Planners, by
contrast, are judged on collision, progress, comfort, or combined driving
scores.  These measures can disagree: a forecast can improve without changing
the selected action, while two forecasts with identical actor-level scores can
make the planner choose differently.

The distinction matters especially for joint forecasts.  A marginal
predictor gives each actor its own possible futures, whereas a joint predictor
describes possible futures for the whole scene.  Each joint mode pairs one AV
future with one traffic-scene future: one mode may pair ``the AV proceeds'' with
``cross traffic yields,'' for example, while another pairs ``the AV waits'' with
``cross traffic proceeds.''  We call this relationship the AV--traffic
association, or pairing; throughout, \emph{marginal} means the forecast for one
actor viewed alone.

Comparing two separately trained models cannot answer whether this pairing is
useful, because their trajectories, probabilities, diversity, training
objectives, and support for planner candidates all differ, and any of those
differences could change what the planner selects.

We therefore ask a narrower question: \emph{if we keep the
predicted trajectories and each actor's probabilities fixed, how does removing
learned AV--traffic pairing affect planner decisions?}  We modify the output
of one joint predictor by pairing every learned AV future with every learned
traffic-scene future and assigning each combination a product-based weight.
This removes the learned relationship between the AV mode and the traffic scene
without changing either side when viewed separately before planner
conditioning; all traffic actors also remain together within each traffic-scene
mode.  The comparison captures both learned-mode pairing removal and the concentration
change it produces under the fixed planner interface.

We call the modified forecast the \emph{control} and compare it with the original
joint forecast in the same candidate planner.  On 1,400 held-out Argoverse 2
scenarios, two training-set sizes, and six seeds per size, we trace the contrast
from candidate costs and ranks through selections and recorded-trajectory
outcomes.  The decision funnel shows where the two forecasts begin to diverge
and how the contrast depends on conditioning.  We
report $\tau=4$~m and the sharper $\tau=1$~m as two conditioning settings.

We contribute (1) a marginal-preserving intervention that removes learned
AV--traffic pairing within one forecast and traces its consequences under a
fixed planner interface; (2) a decision funnel from candidate costs through
recorded-trajectory outcomes; and (3) controlled
pairing-strength, conditioning, geometry, and outcome-horizon sensitivities
across twelve trained runs.

\section{RELATED WORK}

\subsection{Joint Multi-Agent Forecasting}

Scene Transformer uses one attention architecture for both marginal and joint
trajectory prediction~\cite{ngiam2022scene}.  M2I first predicts an influencing
actor and then predicts the responding actor~\cite{sun2022m2i}.  ScePT produces
scene-consistent predictions for planning~\cite{chen2022scept}, and FJMP learns
a directed graph of actor interactions~\cite{rowe2023fjmp}.  MotionDiffuser
models a controllable joint distribution with diffusion~\cite{jiang2023motiondiffuser},
while MotionLM generates interactive futures autoregressively~\cite{seff2023motionlm}.
Konstantinidis et al.\ compare product-weighted recombinations of marginal
modes with jointly trained variants~\cite{konstantinidis2025marginal}.  These
methods construct or learn coherent scene supports.  Our intervention instead
starts with one learned joint forecast and removes its learned-mode AV--traffic
pairing, preserving both learned marginal supports and their probabilities;
architecture-level comparisons change those quantities as well.

\subsection{Prediction Evaluation for Planning}

LookOut connects joint futures to contingency plans~\cite{cui2021lookout}, while
Contingencies from Observations learns a compact contingency policy from
multi-agent behavioral observations~\cite{rhinehart2021contingencies}.  The
Multiple Futures Prediction framework conditions other-agent futures on a
hypothetical ego rollout for planning~\cite{tang2019multiple}.  The
Waymo Open Motion Dataset includes interaction data and joint forecast
metrics~\cite{ettinger2021waymo}.  Planning-aware metrics give more weight to
errors that may affect later decisions~\cite{ivanovic2022planning,philion2020pkl}.
Other studies connect prediction quality to closed-loop behavior and driving
performance~\cite{bouzidi2025closing}.  The nuPlan benchmark evaluates planners
in closed-loop simulation~\cite{karnchanachari2024nuplan}, while Dauner et al.\
show that simple rule-based priors can outperform more elaborate learned
planners in that benchmark~\cite{dauner2023parting}.  Da et al.\ adaptively
combine forecast accuracy and diversity based on scenario criticality to align
evaluation with closed-loop driving performance~\cite{da2026measuring}.  CaAD
uses joint-mode embeddings and planning-oriented feedback to model ego--traffic
dependencies directly~\cite{moon2026caad}.  Our controlled comparison asks a
different question: it keeps actor-level forecast marginals fixed before
planner conditioning and removes learned-mode pairing in one fixed planner.

\section{CONTROLLED PLANNING EVALUATION}

\subsection{Data and Forecast Models}

Recorded futures enter only after candidate selection.  They evaluate the
decision but never influence it.

We use the Argoverse 2 Motion Forecasting Dataset~\cite{wilson2021av2}.  Each
input contains 5~s of actor history and a local high-definition map, and the
model predicts the next 6~s at 10~Hz.  All positions are expressed relative to
the AV; we include up to 48 surrounding actors within 100~m at the last observed
time.  Future data are never used to choose actors, generate candidates, or
select an action.

All learned predictors share a small recurrent actor encoder and transformer
backbone with width 64, four attention heads, and two transformer layers.  The
joint predictor has 120,894 parameters; the separately trained marginal
predictor has 151,774 and decodes a six-mode distribution for
each actor, whereas the joint predictor shares one global mode index and one
scene-level probability head.  Actor inputs represent position, velocity, heading,
type, and observation validity; map inputs represent nearby lanes and
crosswalks.  The marginal model predicts six trajectories and their
probabilities for each actor separately, whereas the joint model predicts six
possible scenes, each with one AV trajectory, one trajectory for every selected
traffic actor, and a probability for the whole scene.  During joint training,
the loss chooses the scene mode that best matches the AV and valid traffic
actors, balancing AV error against average traffic-actor error so scenes with
more actors do not receive more weight.  Mode zero is a fixed constant-velocity
forecast in both models; the other five modes are learned.  We use AdamW with
learning rate $3\times10^{-4}$, batch size 8, at most 50 epochs, and early
stopping.  Matched models are trained on 1,000 and 5,000 scenes using seeds 17,
29, 43, 47, 53, and 59, with checkpoint selection on a separate, fixed 200-scene
validation set.

For the joint predictor, mean-pooled scene features produce six logits, whose
softmax gives $p_m$.  The unit-weight classification term is cross-entropy
against the soft target proportional to $\exp(-e_m/1\,\mathrm{m})$, where $e_m$
is group-balanced scene ADE.  The trajectory term assigns each scene to its
best learned mode among modes 1--5.  Mode zero is exact constant velocity and
receives no trajectory regression, but it remains in the six-way probability
target and loss.  We apply no post-hoc probability calibration.

\subsection{Marginal-Preserving Product Control}

\begin{figure}[t]
    \centering
    \includegraphics[width=\columnwidth,trim=0 38bp 0 0,clip]{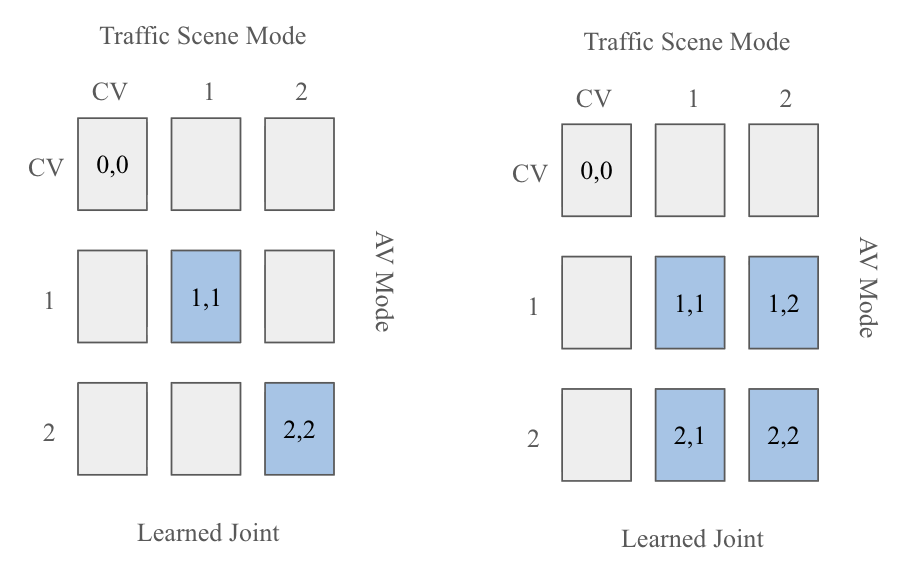}
    \vspace{-1mm}

    {\footnotesize
    \makebox[0.49\columnwidth][c]{Learned joint}\hfill
    \makebox[0.49\columnwidth][c]{Product control}}
    \caption{Learned and product-control AV--traffic pairings, illustrated with
    two learned modes.  The learned forecast retains same-index pairs; the
    control forms all learned-mode combinations.  The evaluated model has one
    constant-velocity pair and five learned modes, giving
    $1+5\times5=26$ control combinations.}
    \label{fig:pairing-comparison}
\end{figure}

Figure~\ref{fig:pairing-comparison} replaces the learned same-index pairs with
their Cartesian product; the fixed constant-velocity pair is unchanged.

Each joint mode $m\in\{0,\ldots,K\}$ contains an AV trajectory $x_m$, a
complete traffic scene $y_m$, and a probability $p_m$, where $K=5$.  The
learned forecast pairs $x_m$ only with $y_m$; the control keeps mode zero as the
fixed constant-velocity pair but forms every combination $(x_i,y_j)$ among the
five learned modes.  Let $L=\sum_{m=1}^{K}p_m$ be their total probability.  The
control assigns each new combination the probability
\begin{equation}
q_{ij}=\frac{p_i p_j}{L},\qquad i,j\in\{1,\ldots,K\}.
\label{eq:control}
\end{equation}
Before planner conditioning, the probability distribution for the AV alone and
for traffic alone stays the same because
\begin{equation}
\sum_{j=1}^{K}q_{ij}=p_i,\qquad
\sum_{i=1}^{K}q_{ij}=p_j.
\label{eq:marginals}
\end{equation}
For a fixed AV future $x_i$, the probabilities of all combinations containing
$x_i$ add back to $p_i$, and the same identity holds for each traffic scene
$y_j$.  The AV forecast and every traffic actor's forecast are therefore
unchanged when viewed separately; keeping all traffic actors together inside
each $y_j$ also preserves their relationships with one another.  The control
generates no new trajectory.  Some AV--traffic combinations may be physically
unlikely, but this is an analysis tool rather than a forecast for deployment.

The planner later gives more weight to AV futures close to candidate $c$.  The
control then has a useful property: within the five learned traffic scenes,
their relative weights do not depend on the candidate; only the total weight of
the learned mixture relative to the fixed constant-velocity pair can change.
Let $k_i(c)=\exp[-d(c,x_i)/\tau]$ and
$A(c)=\sum_{i=1}^{K}p_i k_i(c)$.  Summing over the AV modes gives the traffic
weights
\begin{equation}
\begin{aligned}
\widetilde w_0(c)&=\frac{p_0k_0(c)}{p_0k_0(c)+A(c)},\\
\widetilde w_j(c)&=\frac{A(c)}{p_0k_0(c)+A(c)}\frac{p_j}{L},\quad j\geq1.
\end{aligned}
\label{eq:control_conditioned}
\end{equation}
Equation~(\ref{eq:control_conditioned}) shows that the relative weight of
learned traffic scene $j$ is always $p_j/L$.  The implementation lists all 26
combinations, but only the same six traffic scenes need to be scored.  The
larger list therefore gives the control no selection advantage.  In our runs,
this candidate-independent relative weighting produces a flatter conditioned
control distribution.  The observed concentration change is induced by
removing learned-mode pairing under Eq.~(\ref{eq:conditioning}); the
intervention therefore includes rather than fixes conditioned concentration.
For the six collapsed traffic-scene weights $w_m(c)$, we define the effective
number of modes as
$N_{\mathrm{eff}}(c)=\exp[-\sum_m w_m(c)\log w_m(c)]$.  For reported means, we
average candidates within each scenario, scenarios within each trained run,
and finally the runs themselves.

\subsection{Fixed Candidate Planner}

Each candidate starts at the observed AV state and follows one of up to four
map routes.  We combine seven target forward accelerations, from $-4$ to
$2$~m/s$^2$, with three lateral targets: the lane center and offsets of
$\pm0.75$~m.  We reject trajectories that violate limits on speed, acceleration,
jerk, curvature, lateral acceleration, or map containment, retaining at most
40 candidates per scene and 14.97 on average.  Because Argoverse 2 does not
provide a navigation command for this study, candidates are compared only
within the same route; the forecast does not choose the destination.

The planner cost combines collision, clearance, progress, and comfort with
weights 100, 10, 1, and 1.  Its unweighted components are an indicator of any
horizon overlap; the time mean of
$[\max\{0,(4-r_t)/4\}]^2$, where $r_t$ is the minimum oriented-box clearance;
$-s/50$, where $s$ is route progress in meters; and the comfort term
$\overline{(a_t/4)^2}+\overline{(j_t/6)^2}+
\overline{(a_t^{\mathrm{lat}}/4)^2}$.  The denominators are in m/s$^2$,
m/s$^3$, and m/s$^2$, respectively, and overbars denote time means.  Because
AV2 tracks do not contain object dimensions,
we approximate oriented footprints by type: the AV and vehicles are
$4.8\times2.0$~m, pedestrians $0.8\times0.8$~m, cyclists
$1.8\times0.7$~m, motorcyclists $2.2\times0.9$~m, and buses
$12.0\times2.6$~m.  Collision is rectangle overlap under a separating-axis
test; clearance is a conservative separating-axis gap with a squared penalty
inside 4~m.  For each candidate $c$, the planner updates the scene probabilities
according to how closely the predicted AV path matches the candidate:
\begin{equation}
\widetilde p_m(c)\propto p_m
\exp\!\left[-d(c,x_m)/\tau\right].
\label{eq:conditioning}
\end{equation}
Here, $d$ is the mean distance over 6~s.  We report $\tau=4$~m and $\tau=1$~m,
with other values used in an exploratory sweep.  A closer AV mode receives
more weight.  The planner computes a candidate's cost by averaging the traffic
costs using these updated probabilities.  The joint forecast and the control
use the same formula, costs, candidates, and tie-breaking rule.

We call a candidate \emph{supported} when it is close enough to at least one
predicted AV mode.  For each trained run, we choose the distance threshold on a
separate set of 300 development scenarios.  The target is 90\% scenario
coverage, and the resulting thresholds range from 8.47 to 11.24~m.  If any
candidate on a route is unsupported, both methods use the same conservative
marginal fallback for the whole route.  It rescores every route candidate with
the separately trained marginal predictor: each traffic actor has an
independent mode distribution, one sampled mode persists through the horizon,
and the expectation preserves any-collision and closest-clearance semantics.
No route mixes joint and marginal scores.  The joint forecast and control
contain the same AV trajectories, so they always make the same support and
fallback decisions.  Any selection difference must therefore come from the
removal of learned-mode pairing and the induced concentration change.

\subsection{Recorded-Trajectory Outcomes and Reported Measures}

After the planner selects a candidate, we compare it with recorded future
traffic only at times when future actor states are available.
Recorded-trajectory overlap regret is the candidate's recorded-traffic cost
minus the lowest such cost on the same route, so lower is better; we also
record whether its footprint overlaps a recorded actor.

Many actor tracks end before the full 6~s horizon, so these outcomes are only
partly observed: on average, 72.6\% of actor/time pairs are available.  Recorded
traffic also cannot react to a different AV action.  Only 15 of 1,400 scenarios
have complete 6~s data for every selected traffic actor; the other scenarios
still contribute at observed times.  We therefore treat these as secondary
diagnostics and call the measure ``recorded-trajectory overlap.''

For each route and trained run, we report whether the control changes relative
costs, any candidate rank, the selection before fallback, or the offline
selection.  Relative costs count as changed when the within-route range of
control-minus-joint candidate-cost differences exceeds $10^{-12}$.  A rank
counts as changed when the complete order under deterministic sorting by
(cost, candidate identifier) differs.  For recorded-trajectory outcomes, we first average routes within each scenario and then
subtract the joint result from the control result, so a negative regret
difference favors the control.  Confidence intervals for one trained run use
10,000 scenario-bootstrap samples.  The combined bootstrap resamples both the
six seeds and the same scenario indices across seeds, following the
crossed-data principle of the pigeonhole bootstrap~\cite{owen2007pigeonhole}.
For pooled selection-change intervals, each replicate holds the two training
sizes fixed and recomputes the ratio of changed selections to routes.
Six seeds provide only a limited estimate of seed uncertainty, and every seed shares the
same 1,400 scenarios.  We therefore report the $t$ interval over the six run
effects descriptively rather than treat it as a formal test.  We also repeat
the bootstrap while preserving all six
fixed-stratum counts.

\section{EXPERIMENTAL SETUP}

We selected the final 1,400 scenarios using only observed motion and map
information, before viewing the final predictions or outcomes.  The split uses
a fixed mixture of six observed-only strata: 672 converging intersection scenes
with multiple routes, 28 other converging multiple-route scenes, 56 converging
single-route scenes, 462 nonconverging intersection scenes with multiple routes,
14 other multiple-route scenes, and 168 single-route controls.  It contains 756
scenes where observed paths converge and 1,292 scenes with an observed actor
within 20~m of the AV\@.  We call each combination of training size and seed a
\emph{trained run}; all twelve process every scenario without failure, giving
16,800 scenario/run evaluations and 43,968 route/run evaluations.  Every
matched comparison uses the same candidates, route identifiers, cost settings,
support calibration, and outcome code.

We assign split roles by scenario identifier.  The 1,000-scene training
set is nested within the 5,000-scene set, both selected from the AV2 training
partition; all runs share a separate 200-scene checkpoint-validation set.  The
300 support-development scenes and 1,400 final evaluation scenes come from the
AV2 validation partition.  Before final evaluation, identifier intersections
were checked to confirm that checkpoint validation, support development, and
final evaluation were mutually disjoint and excluded all training scenarios.

\paragraph{Analysis settings}
The original study plan specified $\tau=4$~m.  The first three seeds were
initially evaluated with the evaluator's default setting, $\tau=1$~m, and were
inspected before the seed extension.  The three added seeds are therefore a
post-opened precision extension, not an independent replication.  We then held
$\tau=1$~m fixed for the
three added seeds and evaluated $\tau=4$~m by planner-only rescoring.  Both comparisons remove
learned-mode AV--traffic pairing and include the resulting change in
candidate-conditioned concentration.  The six-seed interval estimates are
therefore descriptive.  The fixed-stratum split emphasizes interaction-rich
scenes rather than the full AV2 driving population.

\paragraph{Evaluation checks}
Within each training size, all six seeds use the same model architecture,
training settings, data partition, candidate planner, and outcome code.  The
added seeds repeat the same pipeline on the same 1,400 scenarios.  Each trained
run completes all 3,664 route comparisons, and the joint forecast and control share candidates,
support decisions, and fallbacks within every comparison.

\paragraph{Analysis reconstruction}
We rebuild the tables from stored per-scenario outputs using 10,000 bootstrap
samples and fixed seed 20290728.  Rebuilding the earlier three-seed tables
changes no shared entry by more than $10^{-9}$.  The sensitivity analyses reuse
stored forecasts; they do not repeat training or inference.

\paragraph{Pairing-strength and conditioning sweep}

We also use the stored forecasts for an exploratory sensitivity analysis.
For learned AV mode $i$ and traffic scene $j$, we define the coupling
\begin{equation}
C_{\lambda}(i,j)=(1-\lambda)p_i\mathbf{1}\{i=j\}
    +\lambda\frac{p_i p_j}{L},\qquad i,j\geq1.
\label{eq:coupling_sweep}
\end{equation}
Mode zero remains fixed.  Thus $\lambda=0$ is the learned pairing,
$\lambda=1$ is the control, and intermediate values preserve both weighted
marginals.  We evaluate
$\lambda\in\{0,0.25,0.5,0.75,1\}$ and
$\tau\in\{0.25,0.5,1,2,4,8,16\}$~m.  Each grid point uses the same six traffic
scenes, candidates, support decisions, costs, and outcomes; no training or
inference is repeated.  We highlight the reported $\tau=4$ and $\tau=1$
settings; intermediate $\lambda$ values and the remaining temperatures
are exploratory.  Their intervals are descriptive and unadjusted.

\paragraph{Geometry and outcome-horizon sensitivities}
Using the same stored forecasts, we uniformly scale the AV and traffic
rectangles used by the cost function by $0.8$, $1.0$, and $1.2$ while holding
candidate generation, map containment, support decisions, and the $\tau=1$
setting fixed.
Separately, we keep the baseline 6~s selections fixed and truncate only the
recorded-traffic collision and clearance calculation to 2 or 4~s; full-plan
progress and comfort remain unchanged.  Neither analysis repeats training or
inference.  Each reproduces the baseline 6~s, unit-footprint endpoint before
moving to the new settings.  These checks probe the sharper $\tau=1$ result.

\section{RESULTS}

\subsection{Conditioning at $\tau=4$}

\begin{figure*}[t]
    \centering
    \includegraphics[width=0.94\textwidth]{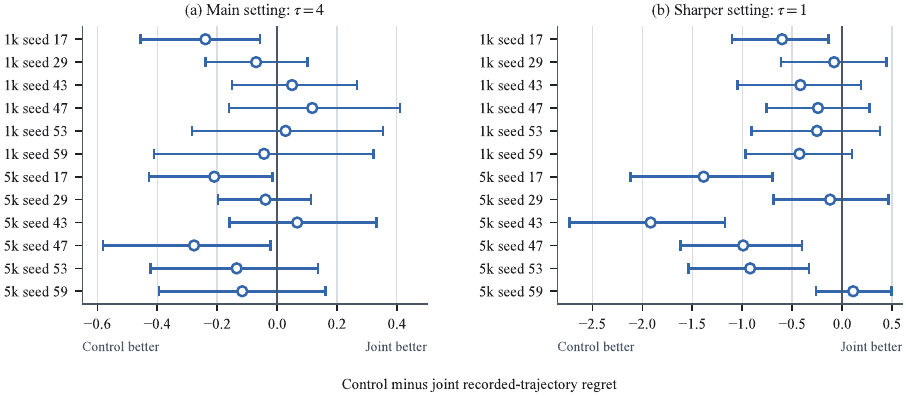}
    \caption{Run-level recorded-trajectory regret at $\tau=4$ and $\tau=1$.
    Negative values favor the product control.  Bars
    are 95\% fixed-run scenario-bootstrap intervals.}
    \label{fig:run-forest}
\end{figure*}

At $\tau=4$, the intervention removes learned-mode pairing, changes
conditioned concentration, and alters 1,422 of 43,968 pre-fallback selections (3.2\%) and 1,312
route-level offline selections (3.0\%; crossed 95\% interval, 2.4--3.6\%).
The offline rates are 3.4\% for the 1,000-scene
models and 2.5\% for the 5,000-scene models.  Mean recorded-trajectory regret
differences are $-0.026$ and $-0.118$ cost units (control minus joint); both
seed-$t$ and crossed intervals span zero (Table~\ref{tab:aggregate}).  Eight of
twelve run estimates favor the control; three fixed-run intervals exclude zero
in that direction, and none favors the joint model (Fig.~\ref{fig:run-forest} and
Table~\ref{tab:run-effects}).

\begin{table}[t]
\caption{Results at Two Conditioning Settings.  Each contrast removes
learned-mode AV--traffic pairing and includes the resulting concentration
change. Selection changes are route-weighted. Regret is control minus joint;
both 95\% intervals apply to regret.}
\label{tab:aggregate}
\centering
\scriptsize
\setlength{\tabcolsep}{1.7pt}
\begin{tabular}{@{}ccccc@{}}
\toprule
Setting & Train & Selection change / & Regret crossed & Regret seed-$t$ \\
& scenes & mean regret $\Delta$ & 95\% interval & 95\% interval \\
\midrule
$\tau=4$ & 1,000 & 3.4\% / $-0.026$ & $[-0.233,0.179]$ & $[-0.156,0.104]$ \\
& 5,000 & 2.5\% / $-0.118$ & $[-0.304,0.058]$ & $[-0.246,0.010]$ \\
\midrule
$\tau=1$ & 1,000 & 8.5\% / $-0.336$ & $[-0.782,0.098]$ & $[-0.528,-0.144]$ \\
& 5,000 & 7.7\% / $-0.870$ & $[-1.590,-0.216]$ & $[-1.671,-0.070]$ \\
\bottomrule
\end{tabular}
\end{table}

\begin{table}[t]
\caption{Run-Level Recorded-Trajectory Regret Effects.  Entries are control
minus joint with fixed-run scenario-bootstrap 95\% intervals, matching
Fig.~\ref{fig:run-forest}; negative values favor the control.}
\label{tab:run-effects}
\centering
\scriptsize
\setlength{\tabcolsep}{1.6pt}
\begin{tabular}{@{}lcc@{}}
\toprule
Run & $\tau=4$ & $\tau=1$ \\
\midrule
1k seed 17 & $-0.239\;[-0.455,-0.057]$ & $-0.601\;[-1.104,-0.133]$ \\
1k seed 29 & $-0.070\;[-0.239,\phantom{-}0.102]$ & $-0.080\;[-0.611,\phantom{-}0.448]$ \\
1k seed 43 & $\phantom{-}0.050\;[-0.150,\phantom{-}0.267]$ & $-0.417\;[-1.049,\phantom{-}0.191]$ \\
1k seed 47 & $\phantom{-}0.118\;[-0.161,\phantom{-}0.411]$ & $-0.241\;[-0.756,\phantom{-}0.275]$ \\
1k seed 53 & $\phantom{-}0.029\;[-0.284,\phantom{-}0.355]$ & $-0.251\;[-0.905,\phantom{-}0.379]$ \\
1k seed 59 & $-0.043\;[-0.411,\phantom{-}0.323]$ & $-0.426\;[-0.965,\phantom{-}0.098]$ \\
5k seed 17 & $-0.209\;[-0.427,-0.015]$ & $-1.387\;[-2.117,-0.696]$ \\
5k seed 29 & $-0.038\;[-0.197,\phantom{-}0.113]$ & $-0.119\;[-0.685,\phantom{-}0.465]$ \\
5k seed 43 & $\phantom{-}0.067\;[-0.158,\phantom{-}0.333]$ & $-1.918\;[-2.731,-1.172]$ \\
5k seed 47 & $-0.277\;[-0.582,-0.021]$ & $-0.990\;[-1.619,-0.403]$ \\
5k seed 53 & $-0.135\;[-0.423,\phantom{-}0.137]$ & $-0.921\;[-1.539,-0.333]$ \\
5k seed 59 & $-0.116\;[-0.395,\phantom{-}0.162]$ & $\phantom{-}0.112\;[-0.259,\phantom{-}0.495]$ \\
\bottomrule
\end{tabular}
\end{table}

\subsection{Sharper Conditioning at $\tau=1$}

\begin{figure}[t]
    \centering
    \includegraphics[width=\columnwidth]{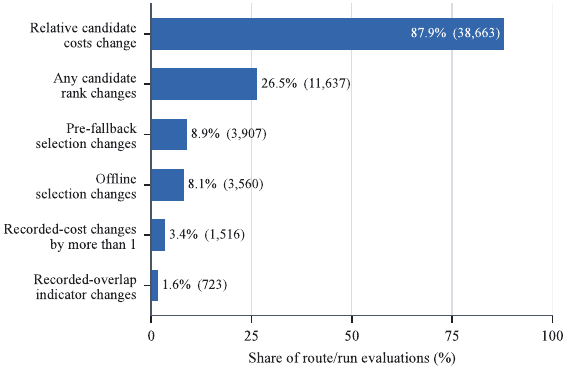}
    \caption{Decision funnel at $\tau=1$, pooled over twelve runs.  The
    intervention removes learned-mode AV--traffic pairing and changes
    conditioned concentration; the same 1,400 scenarios appear in every run.}
    \label{fig:funnel}
\end{figure}

At $\tau=1$, the intervention
changes relative candidate costs in 38,663 of 43,968 route/run evaluations
(87.9\%), at least one rank in 26.5\%, the pre-fallback selection in 8.9\%, and
the offline selection in 8.1\% (Fig.~\ref{fig:funnel}).  Most cost and rank
changes affect only losing candidates and never change the selected action.
The crossed 95\% interval for the 8.1\% rate is 7.1--9.1\%.
The support rule removes 347 of the 3,907 pre-fallback selection changes.

At this sharper setting, the joint forecast averages 2.53 effective traffic
modes after candidate conditioning, compared with 4.66 for the control.
The candidate-conditioned divergence
$D_{\mathrm{KL}}(w_{\mathrm{joint}}\|w_{\mathrm{control}})$ is 0.75--0.98
nats, and total variation is 0.43--0.50 across runs.  This concentration
difference is part of the $\tau=1$ intervention that removes learned-mode pairing.

Eleven of twelve regret estimates favor the control and five fixed-run
intervals exclude zero.  The outcome signal is sparse: 87.6\% of scenario/run
differences are exactly zero, and only 723 of 43,968 route/run overlap
indicators change.  The control avoids 487 of those overlaps and causes 236.
Recorded traffic cannot respond to a different AV action, and many tracks end
early.  We therefore give more weight to the controlled decision counts.
With fixed-stratum resampling, the $\tau=1$ crossed intervals are
$[-0.784,0.107]$ at 1,000 training scenes and $[-1.590,-0.206]$ at 5,000,
essentially unchanged from Table~\ref{tab:aggregate}.

Table~\ref{tab:qualitative-costs} gives one AV2 decision change.  Both
candidates share a route and lateral target; $a$ is target forward acceleration
in m/s$^2$, not a forecast-mode index.  Learned pairing selects the more
aggressive candidate ($a=+2$), whereas product control selects $a=+1$.  In this
example, changing the weights on the same traffic forecasts changes the
planner's selection.

\begin{table*}[t]
\caption{Candidate Costs for One AV2 Decision Change.  Lower is better; bold
indicates the planner selection.  Recorded-traffic cost is an outcome
diagnostic only.}
\label{tab:qualitative-costs}
\centering
\small
\setlength{\tabcolsep}{7pt}
\begin{tabular}{@{}ccccc@{}}
\toprule
Candidate $a$ & Route progress at 6~s & Learned-pairing & Product-control & Recorded-traffic \\
(m/s$^2$) & (m) & cost & cost & cost \\
\midrule
$+2$ & 57.3 & \textbf{0.89} & 4.01 & 1.30 \\
$+1$ & 40.1 & 1.53 & \textbf{2.55} & 0.92 \\
\bottomrule
\end{tabular}
\end{table*}

\subsection{Diagnostics for the Sharper Setting}

The footprint and outcome-horizon rows in
Table~\ref{tab:geometry-horizon} probe the sharper $\tau=1$ result.  Scaling both
footprints from $0.8\times$ to $1.2\times$ changes the pooled offline selection-change rate
from 7.6\% to 8.4\% (8.1\% at baseline), without reversing either mean regret
contrast.  Truncating recorded traffic to 2~s makes both contrasts nearly zero;
the gaps appear later in the horizon, where track attrition is greater.

The temperature rows show how conditioning matters.  The contrast grows at
sharper settings and contracts as the weights flatten; at $\tau\geq8$ its absolute mean is at
most 0.057 cost units.  The $\tau=4$ and $\tau=1$ results differ because
candidate conditioning changes the contrast from learned-mode pairing removal
and its concentration change.  The remaining temperatures and
intermediate coupling values are exploratory.

\begin{figure}[t]
    \centering
    \includegraphics[width=\columnwidth]{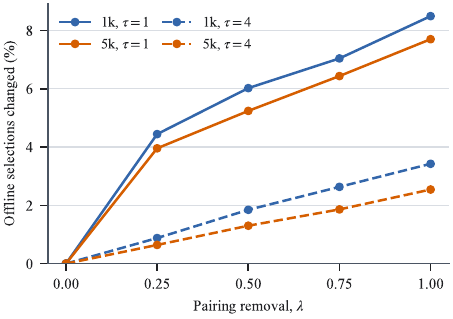}
    \caption{Offline selection changes along the pairing-strength sweep.  Each
    point is route-weighted across 1,400 scenarios and six seeds; the comparator
    is $\lambda=0$ at the same $\tau$.}
    \label{fig:pairing-strength}
\end{figure}

The pairing-strength sweep traces the path between the learned pairing and
product control.  Across both training sizes and both reported $\tau$ values, selection
changes increase monotonically at the five tested $\lambda$ values
(Fig.~\ref{fig:pairing-strength}).  The route-weighted endpoint rates reach
8.5\% and 7.7\% at $\tau=1$, and 3.4\% and 2.5\% at $\tau=4$, for the 1,000-
and 5,000-scene models, respectively.

\begin{table*}[t]
\caption{Conditioning, Geometry, and Outcome Sensitivities.  Values are
control-minus-joint recorded-trajectory regret with unadjusted seed-$t$ 95\%
intervals.  Geometry and horizon rows use $\tau=1$; temperature settings other
than the two marked values are exploratory.}
\label{tab:geometry-horizon}
\centering
\footnotesize
\setlength{\tabcolsep}{8pt}
\begin{tabular}{@{}llcc@{}}
\toprule
Sensitivity & Setting & 1,000 scenes & 5,000 scenes \\
\midrule
Footprint & $0.8\times$ & $-0.788\;[-1.185,-0.391]$ & $-1.016\;[-1.506,-0.525]$ \\
scale & $1.0\times$ & $-0.336\;[-0.528,-0.144]$ & $-0.870\;[-1.671,-0.070]$ \\
& $1.2\times$ & $-0.306\;[-0.672,\phantom{-}0.060]$ & $-0.763\;[-1.325,-0.201]$ \\
\midrule
Outcome & 2~s & $-0.021\;[-0.095,\phantom{-}0.053]$ & $-0.006\;[-0.090,\phantom{-}0.078]$ \\
horizon & 4~s & $-0.230\;[-0.320,-0.140]$ & $-0.554\;[-1.105,-0.004]$ \\
& 6~s & $-0.336\;[-0.528,-0.144]$ & $-0.870\;[-1.671,-0.070]$ \\
\midrule
Condition- & $0.25$ & $-1.293\;[-1.864,-0.722]$ & $-3.505\;[-5.094,-1.916]$ \\
ing $\tau$ (m) & $0.50$ & $-0.844\;[-1.308,-0.380]$ & $-1.924\;[-3.008,-0.841]$ \\
& $1.00$ (reported) & $-0.336\;[-0.528,-0.144]$ & $-0.870\;[-1.671,-0.070]$ \\
& $2.00$ & $-0.121\;[-0.240,-0.002]$ & $-0.262\;[-0.459,-0.065]$ \\
& $4.00$ (reported) & $-0.026\;[-0.156,\phantom{-}0.104]$ & $-0.118\;[-0.246,\phantom{-}0.010]$ \\
& $8.00$ & $-0.001\;[-0.105,\phantom{-}0.103]$ & $-0.057\;[-0.161,\phantom{-}0.048]$ \\
& $16.00$ & $-0.049\;[-0.153,\phantom{-}0.055]$ & $-0.026\;[-0.116,\phantom{-}0.065]$ \\
\bottomrule
\end{tabular}
\end{table*}

\subsection{Approximate Concentration Matching}
\label{sec:matched}

To probe the role of conditioned concentration, we approximately match it.  At
$\tau=1$, the control averages 4.31--5.04 effective modes across runs; the
control's minimum across the matching grid is 3.47--4.12.  For this post-hoc
analysis only, we evaluate $\lambda\in\{0,1\}$ at
$\tau\in\{0.05,0.10,0.25,0.50,1,1.5\}$~m, in 0.25-m steps from 2 to 4.5~m,
and at $\{5,5.5,6,7,8\}$~m.  The joint forecast ranges from 2.10 to 2.78
effective modes at $\tau=1$.  For each run, without using recorded outcomes,
we choose the evaluated joint $\tau$ whose mean effective-mode count is nearest
the control mean at $\tau=1$.  The selected values range from 2.50 to 5.00,
with a mismatch no larger than 0.043 modes.  Each selected grid point remains
fixed during seed/scenario resampling, so the intervals condition on that
selection.

Table~\ref{tab:matched} reports the resulting algebraic decomposition.  The
joint-temperature component is $-0.448$ and $-0.823$ cost units; the remaining
contrasts are $+0.112$ and $-0.047$, and both intervals span zero.  Most of the
$\tau=1$ gap disappears along this path.  The residual remains a mixed
comparison: the arms use different temperatures and match only mean effective
mode count.

\begin{table}[t]
\caption{Algebraic Path at $\tau=1$.  The joint temperature path approximately
matches the control's run-level mean effective mode count.  Components sum
algebraically; brackets are crossed seed/scenario bootstrap 95\% intervals.}
\label{tab:matched}
\centering
\scriptsize
\setlength{\tabcolsep}{1.8pt}
\begin{tabular}{@{}lcc@{}}
\toprule
Component & 1,000 scenes & 5,000 scenes \\
\midrule
Total contrast ($\tau=1$) & $-0.336\;[-0.782,\phantom{-}0.098]$ & $-0.870\;[-1.590,-0.216]$ \\
Joint-$\tau$ path       & $-0.448\;[-0.871,-0.055]$ & $-0.823\;[-1.498,-0.216]$ \\
Residual contrast       & $\phantom{-}0.112\;[-0.094,\phantom{-}0.308]$ & $-0.047\;[-0.297,\phantom{-}0.192]$ \\
\bottomrule
\end{tabular}
\end{table}

\subsection{Actor-Level Forecast Metrics Miss the Difference}

For context, Table~\ref{tab:forecast-sanity} summarizes the models, averaging
first over actors within scenarios and then equally over scenarios and seeds.
The joint forecast and control have identical rows by construction.  Their
top-1 ADE and miss rate are comparable to the separately trained marginal
model, but their minADE is higher and their endpoint diversity is much lower.
The learned joint modes therefore have limited geometric spread, and the
planning result applies to this model--planner instance.

\begin{table}[t]
\caption{Descriptive Forecast Sanity Check.  Errors and weighted endpoint
diversity are in meters; miss uses a 2~m endpoint threshold, and separated modes
use a 2~m separation threshold.  Values are six-seed means.}
\label{tab:forecast-sanity}
\centering
\scriptsize
\setlength{\tabcolsep}{1.4pt}
\resizebox{\columnwidth}{!}{%
\begin{tabular}{@{}rlccccc@{}}
\toprule
Train & Forecast & Top-1 ADE & minADE & Miss (\%) & Diversity & Sep. modes \\
\midrule
1,000 & Joint/control & 1.582 & 1.398 & 35.5 & 0.799 & 1.57 \\
      & Marginal      & 1.742 & 0.969 & 37.1 & 6.893 & 5.03 \\
5,000 & Joint/control & 1.530 & 1.355 & 33.7 & 0.682 & 1.51 \\
      & Marginal      & 1.745 & 0.943 & 35.8 & 6.544 & 4.96 \\
\bottomrule
\end{tabular}%
}
\end{table}

By design, the joint forecast and control have identical top-1 average
displacement error (ADE), minimum ADE (minADE), weighted endpoint diversity,
and every other metric here that scores actors separately.  Even so, the planner
produces different offline selections in 3.0\% of route/run evaluations at
$\tau=4$ and 8.1\% at $\tau=1$.  Actor-by-actor metrics miss both changes.

\section{DISCUSSION}

\subsection{What the Experiment Shows}

At $\tau=4$, removing learned-mode pairing and changing conditioned
concentration alters 3.0\% of route-level offline selections, and both aggregate recorded-outcome
intervals span zero.  At $\tau=1$, it changes 8.1\%; before concentration
matching, the descriptive outcome estimates favor the control.  Approximate
matching attenuates most of this gap, and both residual intervals span zero.
Actor-level metrics are identical in both comparisons.  The main result is
simple: the intervention changes planner decisions, and the size of that change
depends on planner conditioning.  The experiment does not show a
learned-pairing-specific recorded-outcome benefit.

Two mechanisms explain why many cost changes do not reach the final decision.
First, geometrically similar joint modes can describe similar traffic motion
despite having different labels.  Second, the planner selects only the
lowest-cost candidate, so changes among losing candidates do not affect its
choice.  The support fallback plays a smaller role, removing fewer than 10\% of
the selection changes that occur before fallback.

\subsection{Learned-Mode Pairing and Planner Conditioning}

Equation~(\ref{eq:control_conditioned}) means that a candidate cannot change the
relative weights of the control's five learned traffic scenes; it can only
change their total weight relative to the fixed constant-velocity pair.  The
learned joint forecast behaves differently, because a candidate close to one AV
mode increases the weight of the traffic scene paired with that mode.  As
$\tau$ becomes very large, every AV mode receives a similar compatibility
weight and the distinction disappears.  The resulting concentration gap (2.53
against 4.66 effective modes) helps explain why the two forecasts lead to
different planner costs, while the actor-level forecast marginals stay
unchanged before conditioning.

The sweep changes smoothly with $\lambda$, and the contrast becomes small at
large $\tau$, as Eq.~(\ref{eq:coupling_sweep}) predicts.  Approximate matching
also attenuates the $\tau=1$ gap.  Learned-mode pairing and concentration still vary
together in these checks, so the measured contrast belongs to this
forecast--planner combination rather than to the forecast alone.

\subsection{Implication for AV Evaluation}

A joint predictor should be checked at every step between forecasting and
action: actor-level forecast quality, candidate costs, ranks, selected actions,
fallbacks, and outcomes.  Our control removes learned-mode AV--traffic pairing without
training another model or changing actor-level forecast quality before planner
conditioning.  The same test can be applied to larger predictors and other
candidate planners that provide complete scene modes and their probabilities.

\section{LIMITATIONS}

We study one small model family, one candidate generator, and approximate actor
footprints.  The joint predictor has limited geometric mode diversity; a
stronger predictor or a different planner may behave differently.
Each training size has six seeds.  The sensitivity sweeps reuse the same
outcomes, and the matching analysis equalizes only the run-level mean
effective mode count.  Because the two arms use different temperatures, that check
is an algebraic sensitivity analysis rather than a causal decomposition of
learned-mode pairing and concentration.

The control keeps both the relationships among traffic actors and the fixed
constant-velocity pair, removing only the learned relationship between the AV
and traffic modes rather than every relationship between actors.  Finally, many
recorded tracks end early, and recorded traffic cannot react to a different AV
action.  Recorded-trajectory overlap can compare fixed candidates under
recorded traffic, but it cannot establish a causal traffic response or
closed-loop safety.

Argoverse 2 is available under its dataset license.

\section{CONCLUSION}

At $\tau=4$, removing learned-mode pairing and changing conditioned
concentration alters 3.0\% of route-level offline selections, while the recorded-outcome intervals
span zero at both aggregate levels.  At $\tau=1$, the cost and selection changes are larger.  The
unmatched descriptive outcomes favor the control, but approximate concentration
matching attenuates most of that gap and leaves both residual intervals spanning
zero.  Actor-level metrics detect neither case.  For this model--planner pair,
removing learned-mode pairing changes route-level decisions, but the results do not show
a recorded-outcome benefit from learned pairing apart from its accompanying
change in conditioned concentration.  Downstream testing is still needed.

\section*{ACKNOWLEDGMENT}
OpenAI Codex~\cite{openai2026codex} was used to assist with language polishing,
grammar, clarity, and editorial consistency checks throughout the manuscript. The
author reviewed and approved all changes and accepts responsibility for the
technical content, results, citations, and conclusions.

\bibliographystyle{IEEEtran}
\bibliography{references}

\end{document}